\documentclass[10pt,twocolumn,letterpaper]{article}

 \usepackage{wacv}              %

\usepackage{tikz}
\usetikzlibrary{arrows.meta,positioning,calc}

\definecolor{wacvblue}{rgb}{0.21,0.49,0.74}
\usepackage[pagebackref,breaklinks,colorlinks,allcolors=wacvblue]{hyperref}

\def\wacvPaperID{524} %
\def\confName{WACV}
\def\confYear{2027}

\title{Beyond the Frame: Redundancy, Not Trajectory,\\Carries the Privacy Leak in Video Unlearning}

\author{	Kavyansh Tyagi, Paolo Russo\\
Paper ID \wacvPaperID
}

\begin{document}
\maketitle
\begin{abstract}
Class unlearning for video segmentation aims to ensure that a removed
category cannot be reconstructed from a deployed model's outputs. Current
leakage evaluation algorithms inspect one frame at a time, and we show that this understates leakage by a wide margin: an attacker that averages a model's output field over a
two-second clip recovers the forgotten class from every method we evaluate on VSPW and most on Virtual KITTI~2, including models whose per-frame predictions are visually
indistinguishable from a retrain oracle (a model trained from scratch
without the forgotten class). We attribute the recoverable signal to
cross-frame redundancy rather than object trajectory, establish this with a
time-shuffled control that leaves leakage intact, and introduce PACE, an unlearning method that
suppresses redundancy directly using an exactly permutation-invariant pooled
adversarial discriminator and a two-sided temporal consistency objective.
Across VSPW and Virtual KITTI~2, PACE lowers pooled attacker AP to $18.79$
(\texttt{person}) and $12.10$ (\texttt{car}) on VSPW and $9.11$
(\texttt{car}) on Virtual KITTI~2, achieving on the latter the strongest suppression among
verdict-safe methods retaining usable utility, while retaining $19.8$ mIoU
on VSPW \texttt{car} against at most $8.3$ for any baseline achieving
comparable erasure.
\end{abstract}

\section{Introduction}
\label{sec:intro}

\begin{figure*}[t]
    \centering
    \includegraphics[width=0.86\linewidth]{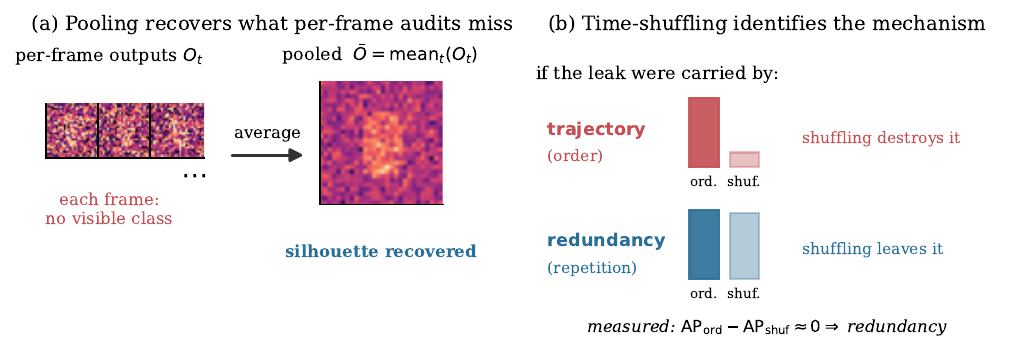}
    \caption{\textbf{The failure mode, and the control that identifies it.}
    (a)~A model can assign negligible probability to the forgotten class in
    every frame and still leak: a weak but temporally consistent residual
    survives averaging while the noise around it does not, and per-frame
    audits cannot observe this quantity.
    (b)~Two mechanisms explain a temporal attacker's advantage. Trajectory
    leakage requires frame order, so permuting frames destroys it; redundancy
    leakage requires only repetition, so permuting frames leaves it intact.
    Scoring both arms separates the two: in every configuration we test,
    the order-dependent component remains at retrain-oracle level.}
    \label{fig:teaser}
\end{figure*}

Regulatory requirements on data deletion~\cite{mantelero2013eu} have made
machine unlearning a practical engineering requirement rather than a
theoretical curiosity~\cite{cao2015towards,bourtoule2021machine,nguyen2022survey}:
make a trained network behave as though a designated class had never
appeared in its training set, at a cost far below full retraining. For
dense prediction tasks the verification step appears straightforward, since
one can query the unlearned model and confirm that the forgotten class no
longer appears in its output.

Video segmentation systems are verified this way, one frame at a time.
Nearly every unlearning method we evaluate appears to erase the forgotten
class at the frame level, yet most retain recoverable signal in their output
fields that an attacker can turn into spatial structure of the forgotten
class from a short video clip. We show that this leakage is large enough to
change the relative ranking of evaluated methods.

The failure originates in a mismatch of observation windows: a per-frame
audit grants the attacker a single image, whereas a deployed video model
exposes a dense sequence of correlated observations of one scene. Suppose an unlearned
model retains a residual preference for the forgotten class too weak to alter
any single frame's arg-max prediction, yet consistent in direction across
frames. Averaging the clip suppresses the independent noise while the residual
accumulates, and the forgotten silhouette becomes recoverable
(Fig.~\ref{fig:teaser}a). No per-frame metric measures this quantity, since by construction
none may observe more than one frame at a time.

We observe the same effect across the evaluated methods: a temporal attacker on the unlearned model's output field over $T{=}16$ frames recovers substantially more of the forgotten class than a per-frame attack does, with the amplification largest precisely where the per-frame audit is most reassuring. At $T{=}3$, frame-adversarial erasure reaches $29.44$ AP per frame, within five points of the retrain oracle, then rises to $45.26$ once pooled, while SalUn~\cite{fan2024salun} moves from $49.27$ to $84.74$ (Table~\ref{tab:t3}). Because these gains differ sharply across methods, per-frame numbers fail even as a relative ordering.

\paragraph{Identifying the carrier.} A temporal attacker outperforming a
per-frame attack admits two explanations, which demand different
defenses. The attacker may read the object's \emph{trajectory}, exploiting
motion cues present in no single frame, which is what temporal leakage
normally denotes; or it may exploit \emph{redundancy}, averaging repeated
observations of an essentially static residual. We separate the two
by applying one random time permutation to both the model's output sequence
and the ground-truth mask sequence, then retraining and rescoring a matched
attacker on the permuted data (Fig.~\ref{fig:teaser}b). Permutation destroys
order while preserving frame content, so the gap
$\mathrm{AP}_{\mathrm{ord}} - \mathrm{AP}_{\mathrm{shuf}}$
isolates the trajectory-borne component. Across three attacker architectures, two
temporal spans, and two forgotten classes, synergy stays within the retrain
oracle's own variation. Redundancy carries the leak: the order-dependent channel
transmits no signal, so an objective enforcing exchangeability over frame
order alone cannot close the gap.

\paragraph{Proposed solution.} This identification is specific
enough to act on. We study a family of adversarial erasure methods, culminating in PACE. PACE-F extends adversarial erasure to individual frames: its discriminator
observes only $O_t$ and reconstructs the forgotten mask $M_t$. PACE-T
(temporal) replaces it with a temporal discriminator and adds a one-sided
bound on the region's frame-to-frame change, which bounds the residual an
averaging attacker can accumulate, and cuts the pooling gain by
$2.5$\,--\,$5\times$ relative to PACE-F under every attacker we test. Reading PACE-T as a variance constraint rather than an order constraint motivates the next step. PACE (pooled) is our proposed method: its adversarial discriminator observes only the time
average $\bar{O} = \frac{1}{T}\sum_t O_t$, making it exactly
permutation-invariant by construction: no redistribution of the residual in
time can satisfy it, so the generator must reduce the pooled residual.
We pair this with a two-sided consistency objective, since the
one-sided form assigns zero penalty to a temporally frozen region, the
configuration most favorable to an averaging attacker.

We repeat the complete pipeline for a second class with sharply different statistics and a second dataset. \texttt{person} is the most frequent class in VSPW~\cite{miao2021vspw} and
deforms as it moves, whereas \texttt{car} is rigid, with nearly linear
motion, and appears in roughly $7.6\times$ fewer eligible training clips; Virtual KITTI~2~\cite{cabon2020vkitti2} is a synthetic driving
benchmark with a different renderer, taxonomy, and motion distribution. The
mechanism and method ranking transfer across both classes and datasets.

\vspace{2pt}
\noindent\textbf{Our core contributions are three-fold:}
\begin{itemize}\itemsep2pt
    \item \textbf{A sound audit for video unlearning.} We show that per-frame
    verification systematically understates recoverable information, and
    introduce a time-shuffled control that decomposes a temporal attacker's
    advantage into redundancy and trajectory terms, establishing redundancy
    as the operative channel across architectures, spans, classes,
    and datasets.
    \item \textbf{An order-invariant adversarial objective.} We introduce
    PACE, whose pooled adversarial discriminator is exactly permutation-invariant by
    construction rather than by training, paired with a two-sided consistency
    term that removes the frozen-region solution of one-sided formulations.
    \item \textbf{Strongest suppression at usable accuracy, on two datasets.}
    PACE attains the lowest pooled leakage of any verdict-safe method that
    retains usable utility, across both VSPW classes and on Virtual KITTI~2;
    on VSPW, PACE and PACE-T are the only methods combining complete
    per-frame erasure with useful utility, where all baselines reaching
    comparable erasure fall to $8.3$ mIoU or below.
\end{itemize}

\section{Related Work}
\label{sec:related}

\paragraph{Approximate Unlearning and Its Verification}
Exact unlearning consists of retraining on the retained data, either directly or through
sharding that limits how much of the model must be rebuilt
~\cite{cao2015towards,bourtoule2021machine}. Because full retraining is costly
at video scale, practical approaches instead modify a trained model so that it
behaves as though the forgotten set were absent. These include gradient ascent
on the forget set and stabilized variants
~\cite{thudi2022unrolling,graves2021amnesiac}; Fisher- and
influence-based parameter scrubbing
~\cite{golatkar2020eternal,golatkar2020forgetting,golatkar2021mixed};
saliency-restricted updates~\cite{fan2024salun}; selective
dampening~\cite{foster2024fast}; teacher--student objectives that repel the
forget set while preserving the retain set
~\cite{kurmanji2023towards,chundawat2023can}; and relabeling with an explicit
substitute target~\cite{tarun2023fast,cha2024learning}.

Most of these methods were developed and evaluated on image classification,
where each sample is observed once, so their verification procedures
inherit a single-sample observation model. Forget-set accuracy alone
is not a sufficient certificate of forgetting~\cite{goel2022evaluating}, while
membership-inference attacks provide a stronger test of residual information
~\cite{shokri2017membership,carlini2022membership}. Weak evaluations can
create a false sense of privacy~\cite{hayes2024inexact}, and the first NeurIPS
unlearning competition showed that method rankings can change with the
evaluation protocol~\cite{triantafillou2024neurips}. These
works strengthen the attacker's inference procedure while keeping its
observation window fixed at one sample. Our setting changes the observation
window itself: a video model exposes a sequence of correlated outputs that
can be jointly analyzed. To our knowledge, prior work has not evaluated
video segmentation unlearning against a temporal pooling attacker, nor
separated the order-dependent and redundancy-based components of the
recovered signal.

\paragraph{Reconstruction from Model Outputs}

Model inversion attacks demonstrate that model outputs can contain information
beyond the nominal prediction task
~\cite{fredrikson2015model,zhang2020secret,he2019model}, and neural networks
can encode attributes that they were not explicitly trained to predict
~\cite{song2019overlearning}. Our attacker is related to this line of work,
but reconstructs the forgotten mask from the segmenter's outputs over the
\emph{retained} classes. In contrast to conventional single-shot inversion,
the signal studied here may be too weak to support reliable recovery from an
individual frame and becomes recoverable only after aggregation across
correlated observations: repeated-query amplification, where the attacker
gains from the redundancy in multiple outputs of the same scene.

\paragraph{Adversarial Suppression and Temporal Modelling}

Suppressing information by training a representation against a discriminator
that attempts to recover it is well established, from domain-adversarial
learning~\cite{ganin2015unsupervised} through censored
representations~\cite{edwards2016censoring} and controllable
invariance~\cite{xie2017controllable}. The same adversarial formulation has
been used for privacy-preserving recognition and anonymization
~\cite{ren2018learning,wu2020privacy,dave2022spact} and for learned
redaction~\cite{orekondy2018connecting}. These approaches suppress the
information accessible to the particular discriminator they employ. Our
setting differs in that the relevant leakage mechanism is first identified
empirically, and the discriminator is then constrained to observe that
quantity---the time-averaged output---making permutation invariance a
property of the attack surface rather than a behavior to be learned. The generator's completion behavior is also related
to context-based inpainting~\cite{pathak2016context,yu2018generative},
although our objective is non-recoverability of a forgotten class rather than
perceptual plausibility.

We build on DeepLabV3~\cite{chen2017rethinking,chen2017deeplab} with a
ResNet-50 backbone~\cite{he2016deep,long2015fully} trained per frame on
VSPW~\cite{miao2021vspw}, following prior observations that competitive
video segmentation can be obtained with per-frame prediction when temporal
consistency is handled separately~\cite{zhu2019improving,kundu2020video};
temporal structure thus enters only through the adversarial discriminator
and the unlearning objective. To test whether our conclusions depend on a
particular temporal architecture, we evaluate a 3D convolutional attacker
~\cite{tran2015learning}, a bidirectional ConvGRU~\cite{cho2014learning}, and
a bidirectional selective state-space model
~\cite{gu2023mamba,gu2022efficiently,li2024videomamba}. The latter has a
full-clip receptive field with $5\times$ fewer parameters than the 3D
attacker, a strong test that the observed absence of an order-dependent
component is not an artifact of insufficient temporal reach.

\section{Methodology}
\label{sec:method}

\begin{figure}[t]
\centering
\begin{tikzpicture}[
  B/.style={draw, rounded corners=2pt, minimum height=0.5cm,
            align=center, font=\scriptsize, inner sep=2pt, fill=white},
  >=Latex, thick
]

\node[font=\tiny\bfseries, gray!70!black] at (-1.55, 1.7) {PACE-F};
\node[B, minimum width=0.95cm]              (Ot)  at (0,    1.7) {$O_t$};
\node[B, fill=gray!12, minimum width=0.85cm](DF)  at (1.75, 1.7) {$D_F$};
\node[B, fill=red!8, minimum width=1.55cm]  (LF)  at (3.75, 1.7) {$\mathcal{L}_{\mathrm{adv}}^{F}$};
\draw[->] (Ot) -- (DF);
\draw[->] (DF) -- (LF);
\node[font=\tiny, gray, align=center] at (0, 2.25) {single frame};

\node[font=\tiny\bfseries, orange!80!black] at (-1.55, 0) {PACE-T};
\node[B, minimum width=0.95cm]                (Os) at (0,    0) {$O_{1:T}$};
\node[B, fill=orange!12, minimum width=0.85cm](DT) at (1.75, 0) {$D_T$};
\node[B, fill=red!8, minimum width=1.9cm, align=center]
     (LT) at (3.9, 0) {$\mathcal{L}_{\mathrm{adv}}^{T}{+}\mathcal{L}_{tc}^{1}$};
\draw[->] (Os) -- (DT);
\draw[->] (DT) -- (LT);
\node[font=\tiny, orange!80!black, align=center] at (0, 0.55) {ordered sequence};

\node[font=\tiny\bfseries, blue!55!black] at (-1.55, -1.7) {PACE};
\node[B, minimum width=0.95cm]              (Ob) at (0,    -1.7) {$\bar{O}$};
\node[B, fill=blue!10, minimum width=0.85cm](DP) at (1.75, -1.7) {$D_P$};
\node[B, fill=red!8, minimum width=1.9cm, align=center]
     (LP) at (3.9, -1.7) {$\mathcal{L}_{\mathrm{adv}}{+}\mathcal{L}_{tc}^{2}$};
\draw[->] (Ob) -- (DP);
\draw[->] (DP) -- (LP);
\node[font=\tiny, blue!55!black, align=center] at (0, -1.15) {time average\\[-1pt]\scriptsize(order-invariant)};

\end{tikzpicture}
\caption{\textbf{The method family.} All three variants train one generator against a discriminator that differs only in what it is permitted to observe: PACE-F's $D_F$ reads a single frame $O_t$; PACE-T's $D_T$ reads the ordered clip $O_{1:T}$, paired with the one-sided bound $\mathcal{L}_{tc}^{1}$ (Eq.~\ref{eq:ltc1}); PACE's $D_P$ reads only the time average $\bar{O}$ (Eq.~\ref{eq:pool}), exactly order-invariant by construction, paired with the two-sided bound $\mathcal{L}_{tc}^{2}$ (Eq.~\ref{eq:ltc2}) that removes the frozen-region solution.}
\label{fig:arch}
\end{figure}
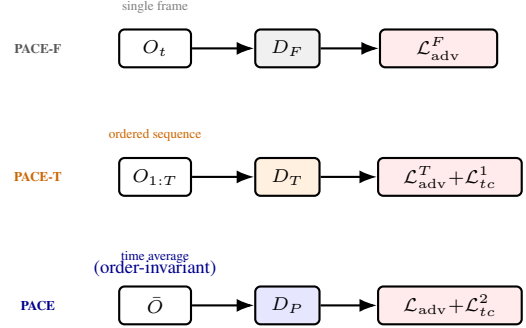

\subsection{Problem Formulation and Threat Model}

Let $f_\theta$ denote a segmenter mapping a frame $I_t \in
\mathbb{R}^{H\times W\times 3}$ to a logit field $f_\theta(I_t) \in
\mathbb{R}^{H\times W\times C}$ over $C$ classes. Unlearning a target class
$c$ produces parameters $\theta'$ such that $c$ is neither predicted nor
recoverable, while accuracy on the retain set $\mathcal{R} = \{1,\dots,C\}
\setminus \{c\}$ is preserved. The supervision target on forgotten pixels is
the dominant-neighbor relabeling $\pi(\cdot)$: each pixel $p$ with
$y_p{=}c$ takes the label of its nearest non-$c$, non-ignore pixel under a
Euclidean distance transform.

Two reference models bracket every measurement. The model \textbf{original} is
trained normally and contains $c$. The model \textbf{oracle} is retrained from
scratch with $\pi$ applied before training begins and defines the privacy
floor. The floor is strictly positive: a model that never observed $c$ can still
encode contextual priors about where $c$-shaped regions tend to occur.

\paragraph{Adversary.}
We grant black-box access to the unlearned model's output field over a clip
of $T$ consecutive frames, together with knowledge of the task, withholding
weights, gradients, and training data: the standard output-access setting
extended to the temporal case. Writing $O_t \in \mathbb{R}^{h\times w\times
|\mathcal{R}|}$ for the retained-class logit field at reduced resolution, the
attacker trains $D_\phi$ to map the clip $O_{1:T}$ to the forgotten mask
sequence $M_{1:T} \in \{0,1\}^{h\times w}$ by minimizing
\begin{equation}
\mathcal{L}_D = \frac{1}{T}\sum_{t=1}^{T}
  \mathrm{BCE}\!\left(D_\phi(O_{1:T})_t,\; M_t\right).
\label{eq:advloss}
\end{equation}

Recovery is scored by average precision (AP) over every pixel in a
held-out set, accumulated through a $512$-bin histogram rather than
subsampling; AP is the primary metric because the positive class occupies a
small fraction of pixels~\cite{davis2006relationship}. We additionally
report AUC, best-threshold IoU, and positive rate as diagnostics.

\subsection{Decomposing the Temporal Advantage}
\label{sec:decomp}

A temporal attacker outperforming a per-frame attacker does not by itself
identify the mechanism: redundancy predicts a variance-reduction effect of
averaging $T$ noisy estimates of a temporally correlated mask, independent of
frame order, whereas trajectory leakage is information carried by the motion
path itself and is order-dependent. We construct a control to separate the
two. Let $\sigma$ be a
permutation of $\{1,\dots,T\}$ drawn independently per clip and applied
jointly to the output and mask sequences, giving $O_{\sigma(1:T)}$ and
$M_{\sigma(1:T)}$; training and evaluating a matched attacker on these
clips removes frame order while preserving per-frame content and
cross-frame redundancy, and we therefore define
\begin{align}
\mathrm{synergy} &= \mathrm{AP}_{\mathrm{ord}}
                   - \mathrm{AP}_{\mathrm{shuf}},
\label{eq:synergy}\\
\Delta_{\mathrm{red}} &= \mathrm{AP}_{\mathrm{shuf}}
                        - \mathrm{AP}_{\mathrm{frame}},
\label{eq:redundancy}
\end{align}
where $\mathrm{AP}_{\mathrm{frame}}$ is a per-frame attacker's score on the
same model; Eq.~\ref{eq:redundancy} measures the recovery available from
cross-frame aggregation, Eq.~\ref{eq:synergy} the order-dependent component.
Reporting both lets the objective target the mechanism the evaluation
identifies.

\paragraph{Protocol controls.}
Three requirements make the decomposition meaningful. Evaluation content must
not vary with $T$: we draw a fixed anchor set of
$(\mathrm{video},\mathrm{frame})$ pairs once and form clips of every
length as windows around the same anchors. The attacker's optimization budget
must remain fixed across $T$: gradient accumulation fixes both optimizer
steps and effective clip batch size, independently of physical batch memory. Finally, scoring must be calibration-free,
since a fixed-threshold IoU can report zero for an attacker whose ranking is
informative but whose sigmoid outputs never cross the threshold.
Section~\ref{sec:ablations} quantifies the effect of each control.

\subsection{Adversarial Erasure Objectives}

All variants train the generator, initialized from \textbf{original}, with
\begin{equation}
\mathcal{L}_G = \lambda_r \mathcal{L}_r
              + \lambda_{\mathrm{adv}} \mathcal{L}_{\mathrm{adv}}
              + \lambda_{tc} \mathcal{L}_{tc},
\label{eq:objective}
\end{equation}
alternating with discriminator updates in the usual minimax
fashion~\cite{goodfellow2014generative}. The retention term
\begin{equation}
\mathcal{L}_r = \mathrm{CE}\!\left(G(I), \pi(y)\right)
   + \beta\, \mathrm{KL}\!\left(G(I)\,\|\,f_\theta(I)\right)\big|_{\mathcal{R}}
\label{eq:lr}
\end{equation}
drives forgotten pixels toward their re-labeled targets while anchoring
retained pixels to the frozen original through a distillation
penalty~\cite{hinton2015distilling}. It is applied on ordinary frames to
limit collateral damage. The variants differ in
$\mathcal{L}_{\mathrm{adv}}$ and $\mathcal{L}_{tc}$, and therefore in the
information they can suppress (Fig.~\ref{fig:arch}).

\paragraph{PACE-F.}
Frame adversarial erasure uses a per-frame discriminator $D_F$ to reconstruct
$M_t$ from $O_t$ alone. It therefore makes each frame individually difficult
to attack, which Sec.~\ref{sec:results} shows is not equivalent to making the
entire clip difficult to attack.

\paragraph{PACE-T.}
Temporal adversarial erasure replaces $D_F$ with a temporal discriminator
over the ordered sequence and adds a one-sided bound on the region's motion,
\begin{equation}
\mathcal{L}_{tc}^{1} = \mathbb{E}\big[\,
  \mathrm{relu}\!\left(\Delta O - \rho\right)\big],
\quad \Delta O = \lVert O_{t+1} - O_t \rVert,
\label{eq:ltc1}
\end{equation}
where the reference $\rho$ is a local motion budget estimated over a dilated
ring surrounding the region and floored by a global estimate. The penalty
therefore activates only when the region changes more than its surroundings.

\subsection{PACE}
\label{sec:pace}

Two changes follow from the decomposition, each addressing a different
failure mode of the objective.

\paragraph{Order-invariant pooled discriminator.}
We introduce a pooled discriminator $D_P$ that observes only
\begin{equation}
\bar{O} = \frac{1}{T}\sum_{t=1}^{T} O_t
\in \mathbb{R}^{h\times w\times|\mathcal{R}|},
\label{eq:pool}
\end{equation}
after temporal averaging and before any learned parameters are applied, and
predicts a single clip-level mask $\bar{M} = \bigvee_t M_t$. Because
averaging precedes the discriminator,
$D_P(\overline{O_{\sigma(1:T)}}) = D_P(\bar{O})$ for every permutation $\sigma$:
the invariance is exact and does not depend on learning.
Numerically, the maximum output deviation under permutation is
$4\times10^{-6}$, at the level of floating-point noise, and
$\partial D_P/\partial O_t$ is uniform over $t$, matching the closed-form
backward pass of the mean.

Redistributing a residual across frames cannot change the pooled
discriminator's input, so the generator must reduce the residual that
survives pooling---the quantity measured by Eq.~\ref{eq:redundancy}. We retain the temporal reconstruction discriminator
$D_T$ alongside $D_P$, giving
\begin{equation}
\mathcal{L}_{\mathrm{adv}} = \underbrace{\mathrm{BCE}\!\left(D_T(O_{1:T}),
  \mathbf{0}\right)}_{\text{temporal reconstruction}}
  + \lambda_{p}\underbrace{\mathrm{BCE}\!\left(D_P(\bar{O}),
  \mathbf{0}\right)}_{\text{pooled reconstruction}} .
\label{eq:ladv}
\end{equation}

\paragraph{Two-sided consistency bound.}
Equation~\ref{eq:ltc1} admits a degenerate solution. A temporally frozen region
has $\Delta O = 0$ and therefore incurs zero penalty, yet such a frozen
residual is particularly favorable to an averaging attacker because
repetition is what pooling exploits. We replace the one-sided form with
\begin{equation}
\mathcal{L}_{tc}^{2} =
\mathbb{E}\big[\left(\Delta O - \rho\right)^{2}\big],
\label{eq:ltc2}
\end{equation}
which penalizes both deficient and excessive motion.

\paragraph{Detached reference.}
We compute $\rho$ without gradient. If $\rho$ were differentiable, the
optimizer could reduce Eq.~\ref{eq:ltc2} by increasing the reference, for
example by adding motion to the surrounding ring rather than reducing the
region's own residual; detaching $\rho$ and flooring it with a global
estimate removes this path, and the same holds for any adversarial penalty
defined relative to a learned reference.

\paragraph{Implementation.}
All variants train on clips of length $3$ at batch size $2$, with
$\lambda_r = \lambda_{\mathrm{adv}} = 1.0$, $\lambda_{tc} = 0.5$, and
$\lambda_p = 1.0$. PACE adds $3.8\%$ wall-clock time over PACE-T
($853.5$ vs.~$822.2$ minutes on one GPU), comparable to the $3.7\%$
overhead of the ablated order term (Sec.~\ref{sec:ablations}). Every attacker is trained from
scratch for each evaluated model and is never shared.

\section{Quantitative and Qualitative Results}
\label{sec:results}

\paragraph{Setup.} Our main study trains on VSPW~\cite{miao2021vspw} ($124$
classes, $2806$ train / $343$ val videos at $15$\,fps), with \texttt{person}
and \texttt{car} as forgotten classes; we additionally evaluate on Virtual
KITTI~2~\cite{cabon2020vkitti2}, a densely annotated synthetic driving
dataset, with \texttt{car} as the forgotten class there too
(\S\ref{sec:vkitti}). Unless stated otherwise, leakage uses the
full-clip-reach selective state-space attacker at $T{=}16$, stride $2$
($2.1$\,s), trained for $500$ optimizer steps at effective batch $4$, with
two training runs per arm scored over $800$ held-out clips; utility is
retain-set mIoU over all non-forgotten classes, and every cache is rebuilt
per class and dataset.

\subsection{Effect of Temporal Pooling on Leakage}

Table~\ref{tab:t3} reports all three arms at $T{=}3$ with the 3D convolutional
attacker, together with the quantities of Eqs.~\ref{eq:synergy}
and~\ref{eq:redundancy}.

\begin{table}[t]
\centering\small
\setlength{\tabcolsep}{4pt}
\begin{tabular}{lccccc}
\toprule
Model & frame & shuf. & ord. & $\Delta_{\mathrm{red}}$ & synergy \\
\midrule
oracle (floor) & 24.73 & 26.77 & 26.99 & $+2.04$ & $+0.22$ \\
original & 91.56 & 94.94 & 94.94 & $+3.38$ & $+0.00$ \\
PACE-F & 29.44 & 45.35 & 45.26 & $\mathbf{+15.91}$ & $-0.09$ \\
PACE-T & 16.42 & 22.62 & 21.26 & $+6.20$ & $-1.36$ \\
SalUn~\cite{fan2024salun} & 49.27 & 83.54 & 84.74 & $\mathbf{+34.27}$ & $+1.20$ \\
\bottomrule
\end{tabular}
\caption{\textbf{Adversary AP by arm} at $T{=}3$. The pooling gain
$\Delta_{\mathrm{red}}$ is largest for the methods that appear most private under a per-frame audit, while synergy remains small relative to the corresponding redundancy gain.}
\label{tab:t3}
\end{table}

The pooling gain concentrates where per-frame evidence is most reassuring.
SalUn sits at $49.27$ AP per frame and reaches $84.74$ pooled, recovering
essentially all of the undefended model's leakage; the undefended model,
already maximally leaky per frame, gains only $3.38$, and the oracle only
$2.04$. PACE-F is the sharpest case: at $29.44$ AP per frame it lies within
five points of the retrain oracle, which a per-frame audit treats as
near-certification, and pools to $45.26$, clearly above the floor: this is
amplification acting on an existing residual, since suppressing amplitude
without removing temporal consistency preserves exactly the structure
pooling exploits.

\subsection{Redundancy versus Trajectory}

We rule out an attacker too myopic to see trajectory, and a clip too short to contain one, by rerunning the control with the strongest attacker at $T{=}16$ across two spans on the most leakage-rich model available. The undefended model records synergy of $-0.01$ at $1.1$\,s and $+0.46$ at $2.1$\,s, both within the oracle's band of $-0.43$ to $+0.86$ under maximally favorable conditions. Three configurations differing in clip length, architecture, and span all place synergy at zero, establishing redundancy as the channel a defense must address at the horizons tested.

\subsection{Suppressing Redundancy via Amplitude Control}

Table~\ref{tab:redundancy} isolates $\Delta_{\mathrm{red}}$ across all three
attacker architectures.

\begin{table}[t]
\centering\small
\setlength{\tabcolsep}{5pt}
\begin{tabular}{lccc}
\toprule
Model & 3D CNN & ConvGRU & Sel.\ SSM \\
\midrule
oracle (floor) & $-0.99$ & $+2.02$ & $+5.49$ \\
original & $+2.09$ & $+3.24$ & $+1.63$ \\
PACE-F & $+12.69$ & $+19.21$ & $+17.49$ \\
PACE-T & $\mathbf{+2.53}$ & $\mathbf{+5.43}$ & $\mathbf{+7.02}$ \\
\bottomrule
\end{tabular}
\caption{\textbf{Pooling gain $\Delta_{\mathrm{red}}$} at $T{=}16$ under three
attacker architectures. PACE-T reduces the gain by $2.5$ to $5\times$ relative to
PACE-F under every attacker, and its residual tracks the irreducible floor as
attacker strength increases.}
\label{tab:redundancy}
\end{table}

PACE-T reduces the pooling gain by factors of $0.20$, $0.28$, and $0.40$ relative
to PACE-F across the three architectures, so the effect is a property of the
objective rather than a weak attacker. The floor supplies the sharper
reading. A model that never observed the class still gains from pooling, and
that gain grows with attacker strength from $-0.99$ to $+2.02$ to $+5.49$. Under
the strongest attacker PACE-T's gain of $+7.02$ tracks the floor's own $+5.49$
while PACE-F's $+17.49$ is independent of it, so PACE-T approaches the irreducible
component whereas PACE-F leaves a leak on top of it.

\subsection{Leak/Safe Verdict Criterion}
\label{sec:verdict}

Tables~\ref{tab:person} and~\ref{tab:car} add a binary \textbf{Verdict}:
\textsc{leak} when ordered AP exceeds the oracle floor's mean by more than
two trial standard deviations ($\mu_{\text{floor}}{+}2\sigma_{\text{floor}}$),
\textsc{safe} otherwise---statistically indistinguishable from retraining
along the pooled channel. The threshold is computed once per class from the floor's own
spread, not tuned per method.

\subsection{Main Results: Person Class}
\label{sec:person}

\begin{table}[t]
\centering\scriptsize
\setlength{\tabcolsep}{3pt}
\renewcommand{\arraystretch}{0.9}
\begin{tabular}{lcccc}
\toprule
Model & retain mIoU $\uparrow$ & person IoU & pooled AP $\downarrow$ & Verdict \\
\midrule
oracle (floor) & 30.8 & 0.0 & 30.82 \small{$\pm$ 0.14} & floor \\
original & 33.5 & 81.2 & 93.82 \small{$\pm$ 0.77} & \textsc{leak} \\
\midrule
FT~\cite{golatkar2020eternal} & 17.4 & 0.0 & 48.48 \small{$\pm$ 1.71} & \textsc{leak} \\
NegGrad~\cite{thudi2022unrolling} & 0.1 & 0.1 & 88.70 \small{$\pm$ 0.39} & \textsc{leak} \\
NegGrad+~\cite{kurmanji2023towards} & 4.9 & 0.0 & 68.14 \small{$\pm$ 1.28} & \textsc{leak} \\
RL~\cite{graves2021amnesiac} & 19.0 & 0.0 & 86.08 \small{$\pm$ 0.40} & \textsc{leak} \\
SalUn~\cite{fan2024salun} & 15.1 & 0.0 & 85.68 \small{$\pm$ 0.35} & \textsc{leak} \\
SSD~\cite{foster2024fast} & 0.4 & 0.0 & 25.57 \small{$\pm$ 1.07} & safe \\
SCRUB~\cite{kurmanji2023towards} & 31.2 & 36.7 & 88.60 \small{$\pm$ 0.35} & \textsc{leak} \\
DELETE~\cite{cha2024learning} & 17.1 & 0.0 & 70.59 \small{$\pm$ 3.57} & \textsc{leak} \\
\midrule
PACE-F & 24.2 & 8.3 & 46.92 \small{$\pm$ 0.38} & \textsc{leak} \\
PACE-T & 22.5 & 1.4 & 21.10 \small{$\pm$ 0.16} & safe \\
\textbf{PACE (ours)} & 21.2 & 0.8 & \textbf{18.79} \small{$\pm$ 1.23} & \textbf{safe} \\
\bottomrule
\end{tabular}
\caption{\textbf{Person class}, all thirteen models under one protocol,
threshold $\mu_{\text{floor}}{+}2\sigma_{\text{floor}}$ from oracle floor. SSD is
the only baseline scoring \textsc{safe}, and it does so by destroying the
model (retain mIoU $0.4$); SCRUB is the one baseline that barely erases the
class at all (person IoU $36.7$) and leaks accordingly. PACE-F, PACE-T and PACE are
the only \textsc{leak}/safe/safe sequence that comes from one continuously
improved objective rather than three unrelated methods, and PACE is the only
method in the table combining a safe verdict, retain mIoU above $21$, and
the lowest pooled AP.}
\label{tab:person}
\end{table}

Eight of ten published baselines leak, and the two exceptions are not the methods a practitioner would want: SSD is safe because retain mIoU has collapsed to $0.4$, not because it forgot selectively. PACE-F, PACE-T and PACE trace one design lineage from \textsc{leak} to safe to a tighter, higher-utility safe, and PACE is the only method that is verdict-safe, has the lowest pooled AP, and retains mIoU above $21$. We return to why PACE-F leaks and PACE-T does not in \S\ref{sec:ablations}.

\subsection{Generalization to the Car Class}
\label{sec:car}

\begin{table}[t]
\centering\scriptsize
\setlength{\tabcolsep}{3pt}
\renewcommand{\arraystretch}{0.9}
\begin{tabular}{lcccc}
\toprule
Model & retain mIoU $\uparrow$ & car IoU & pooled AP $\downarrow$ & Verdict \\
\midrule
oracle (floor) & 31.6 & 0.0 & 18.10 \small{$\pm$ 5.60} & floor \\
original & 31.9 & 33.5 & 80.45 \small{$\pm$ 1.62} & \textsc{leak} \\
\midrule
FT~\cite{golatkar2020eternal} & 6.6 & 0.0 & 32.55 \small{$\pm$ 3.21} & \textsc{leak} \\
NegGrad~\cite{thudi2022unrolling} & 0.1 & 0.0 & 6.97 \small{$\pm$ 0.27} & safe \\
NegGrad+~\cite{kurmanji2023towards} & 3.6 & 0.0 & 16.79 \small{$\pm$ 2.99} & safe \\
RL~\cite{graves2021amnesiac} & 3.7 & 0.0 & 57.61 \small{$\pm$ 4.23} & \textsc{leak} \\
SalUn~\cite{fan2024salun} & 4.5 & 0.0 & 62.85 \small{$\pm$ 0.03} & \textsc{leak} \\
SSD~\cite{foster2024fast} & 0.0 & 0.0 & 7.48 \small{$\pm$ 0.34} & safe \\
SCRUB~\cite{kurmanji2023towards} & 27.5 & 20.3 & 54.41 \small{$\pm$ 2.04} & \textsc{leak} \\
DELETE~\cite{cha2024learning} & 8.3 & 0.0 & 69.57 \small{$\pm$ 1.76} & \textsc{leak} \\
\midrule
PACE-T & 20.2 & 0.0 & 13.93 \small{$\pm$ 0.37} & safe \\
\textbf{PACE (ours)} & 19.8 & 0.0 & \textbf{12.10} \small{$\pm$ 1.81} & \textbf{safe} \\
\bottomrule
\end{tabular}
\caption{\textbf{Car class}, all twelve models under one protocol, threshold
$\mu_{\text{floor}}{+}2\sigma_{\text{floor}}$ AP from oracle floor. Three baselines are
verdict-safe (NegGrad, NegGrad+, SSD); all three reach it with retain mIoU
$\leq 3.6$, i.e.\ by damaging the model rather than by selectively forgetting
\texttt{car}. PACE-T and PACE are the only methods that are simultaneously
verdict-safe and retain mIoU above $8.3$.}
\label{tab:car}
\end{table}

\begin{figure}[t]
    \centering
    \includegraphics[width=0.76\linewidth]{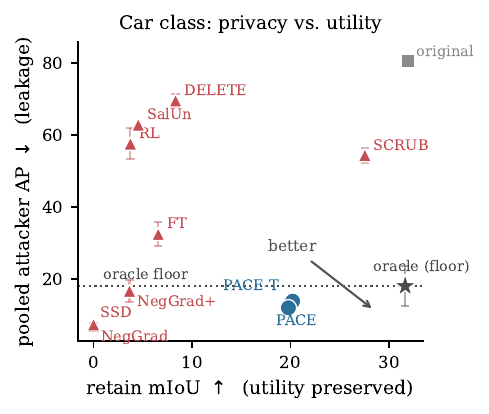}
    \caption{\textbf{Car class, privacy against utility.} Error bars give trial
    spread. Low leakage carries meaning only at usable utility, and the two
    least leaky baselines sit at the left edge because the model has been
    degraded. PACE and PACE-T occupy the useful corner.}
    \label{fig:frontier}
\end{figure}

Tables~\ref{tab:person} and~\ref{tab:car}, with Fig.~\ref{fig:frontier}, make
the same point under different class statistics. A safe verdict is cheap to
obtain by damaging the model: three car baselines and one person baseline
reach it that way, at retain mIoU $\leq 3.6$ and $0.4$. PACE-T
and PACE are, on both classes, the only verdict-safe methods retaining most
non-target accuracy ($20.2$/$19.8$ mIoU on \texttt{car} vs.\ $\leq 8.3$ for
comparable baselines, $22.5$/$21.2$ on \texttt{person} vs.\ SSD's $0.4$), and
PACE attains the lowest AP among them: $12.10$ on \texttt{car} (within the
floor's own $5.60$ spread, so comparable) and $18.79$ on \texttt{person}. SCRUB shows the
opposite failure on both: it preserves utility ($31.2$/$27.5$ mIoU) while
barely erasing the target (IoU $36.7$/$20.3$), leaking from not having
unlearned rather than from redundancy. The transfer to a rigid, rarer class
supports the mechanism over a class-specific artefact.

\subsection{Generalization to Virtual KITTI~2}
\label{sec:vkitti}

\begin{table}[t]
\centering\scriptsize
\setlength{\tabcolsep}{3pt}
\renewcommand{\arraystretch}{0.9}
\begin{tabular}{lcccc}
\toprule
Model & retain mIoU $\uparrow$ & car IoU & pooled AP $\downarrow$ & Verdict \\
\midrule
oracle (floor) & 45.0 & 0.0 & 50.71 \small{$\pm$ 0.71} & floor \\
original & 50.1 & 82.8 & 88.67 \small{$\pm$ 1.27} & \textsc{leak} \\
\midrule
FT~\cite{golatkar2020eternal} & 30.9 & 0.0 & 41.43 \small{$\pm$ 2.32} & safe \\
NegGrad~\cite{thudi2022unrolling} & 1.8 & 0.0 & 9.67 \small{$\pm$ 1.41} & safe \\
NegGrad+~\cite{kurmanji2023towards} & 4.9 & 0.0 & 33.57 \small{$\pm$ 6.40} & safe \\
RL~\cite{graves2021amnesiac} & 31.7 & 0.0 & 89.17 \small{$\pm$ 1.08} & \textsc{leak} \\
SalUn~\cite{fan2024salun} & 32.2 & 0.0 & 92.42 \small{$\pm$ 0.49} & \textsc{leak} \\
SSD~\cite{foster2024fast} & 1.8 & 0.0 & 6.87 \small{$\pm$ 0.19} & safe \\
SCRUB~\cite{kurmanji2023towards} & 44.7 & 10.9 & 95.66 \small{$\pm$ 0.95} & \textsc{leak} \\
DELETE~\cite{cha2024learning} & 32.2 & 0.0 & 79.56 \small{$\pm$ 0.86} & \textsc{leak} \\
\midrule
PACE-T & 21.1 & 2.2 & 17.12 \small{$\pm$ 2.24} & safe \\
\textbf{PACE (ours)} & 7.6 & 0.0 & \textbf{9.11} \small{$\pm$ 0.79} & \textbf{safe} \\
\bottomrule
\end{tabular}
\caption{\textbf{Virtual KITTI~2, car class}, all twelve models under one
protocol, threshold $\mu_{\text{floor}}{+}2\sigma_{\text{floor}}$ AP from oracle floor.
FT is verdict-safe here while retaining $30.9$ mIoU, substantially lower pooled AP than the strongest-leaking baselines in this setting; its synergy of $+10.24$ is also
the single clearly non-zero value across every model and dataset we test.
PACE attains the lowest AP of any safe method but at a lower retain mIoU
($7.6$) than PACE-T ($21.1$).}
\label{tab:vkitti}
\end{table}

Table~\ref{tab:vkitti} repeats the protocol on a second dataset, renderer,
and domain with rigid, fast-moving targets. The pattern from VSPW survives:
four of eight baselines leak, SCRUB again preserves utility ($44.7$ mIoU)
while barely erasing the class (car IoU $10.9$), and PACE-T is again
verdict-safe with the best utility among safe, genuinely-forgetting methods.
Two findings are new, and we report both explicitly. FT
is verdict-safe here while it leaks on both VSPW classes; its synergy of
$+10.24$ is also the only clearly non-zero value across all thirty-six
model-dataset combinations we evaluate, so we treat it as an isolated
exception, not evidence against the decomposition of Sec.~\ref{sec:decomp}. PACE's utility here ($7.6$ mIoU) falls well below
PACE-T's ($21.1$); PACE still reaches the lowest AP of any safe method
($9.11$), but PACE-T is the better operating point on this dataset by
utility, a cost we attribute to the pooled objective interacting with
Virtual KITTI~2's smaller, four-scene training set rather than to pooling
itself: the variants differ only in that discriminator.

\subsection{Qualitative Results}

\begin{figure*}[t]
    \centering
    \includegraphics[width=0.95\linewidth]{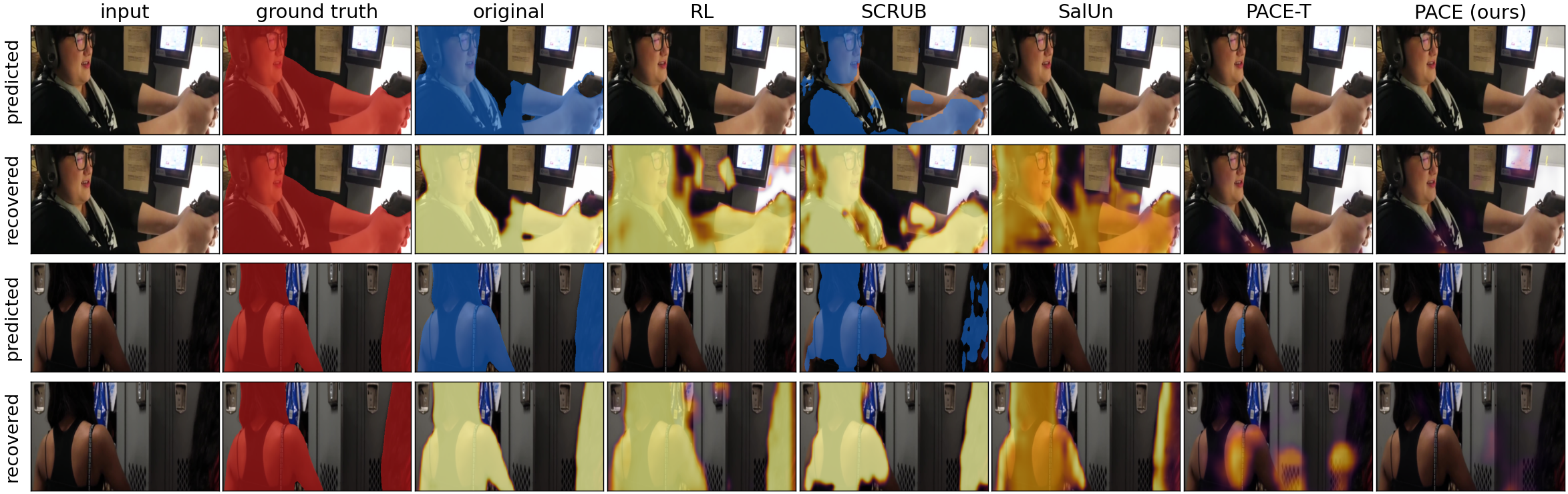}
    \caption{Two \texttt{person} examples: the model's own per-frame
    prediction (predicted row) paired with the pooled attacker's
    reconstruction of the same outputs (recovered row). Red marks the
    ground-truth class, blue the predicted class; brightness in the recovered
    rows is the attacker's probability of the forgotten class. RL and SalUn
    appear to erase the class per frame, yet the pooled attacker recovers a
    clear silhouette (Tab.~\ref{tab:person}); Original and SCRUB already leak
    visibly in their own predictions, while PACE-T and PACE yield
    substantially weaker silhouettes. Attackers use the protocol of
    Tab.~\ref{tab:person} with a reduced single-run budget.}
    \label{fig:attack2}
\end{figure*}

Figure~\ref{fig:attack2} presents two qualitative \texttt{person} examples:
every unlearned model removes the target convincingly per frame, despite
large differences in pooled attack AP (PACE-F: $46.92$; PACE-T: $21.10$;
PACE: $18.79$), illustrating the gap between visual per-frame erasure and
recoverability under the pooled attacker.

\section{Ablation Studies}
\label{sec:ablations}

We isolate one component at a time, holding every other setting fixed at the
configuration of Sec.~\ref{sec:results}. Objective ablations use
\texttt{person}, while attacker and protocol ablations apply to all models.

\subsection{Pooled Adversary and Two-Sided Bound}

Reading Tab.~\ref{tab:person} as a sequence isolates the two changes defining
PACE. PACE-T, which adds a temporal discriminator and the one-sided bound of
Eq.~\ref{eq:ltc1}, reduces pooled leakage from $46.92$ to $21.10$ AP,
identifying amplitude control over the residual as the dominant factor. PACE,
with the pooled discriminator of Eq.~\ref{eq:pool} and the two-sided bound of
Eq.~\ref{eq:ltc2}, lowers it further to $18.79$ AP and tightens the trial
spread to $1.23$.
Both changes move in the direction predicted by Sec.~\ref{sec:decomp}, since
each constrains the pooled residual rather than a proxy for it.

\subsection{Order Adversary and Motion Reference}

PACE-TO adds a discriminator distinguishing true from permuted frame order,
pushing the generator toward exchangeability; since the order channel
measures as zero (Sec.~\ref{sec:results}), it should add little. We also
test PACE-TL, which uses a local motion reference over a dilated ring
instead of PACE-T's global one. Against a same-run PACE-T checkpoint
($21.28 \pm 2.34$ AP, a $0.18$-point gap from Tab.~\ref{tab:person}'s
independently retrained PACE-T), PACE-TL scores $24.81 \pm 0.36$ and
PACE-TO scores $22.70 \pm 2.00$: the order term's $2.11$-point improvement
over PACE-TL is close to the $2.00$-point trial spread, at a $3.7\%$
extra training cost.
The pooled discriminator gets the same invariance by construction at lower
cost, so PACE keeps it and drops the order term. PACE-T's global reference
($21.28$ AP) beats the local one ($24.81$), so PACE keeps the global
estimate, though PACE-TL's tighter spread ($0.36$) suggests a scene-adaptive
reference paired with the pooled discriminator is worth exploring.

\subsection{Adversary Architecture and Temporal Reach}

Conclusions about which channel carries information depend on the attacker
observing it, so we measure receptive field directly by perturbing one
frame's field and recording which outputs change. The 3D
convolutional
attacker~\cite{tran2015learning} at $0.44$M parameters propagates information
only $\pm 3$ frames, observing $7$ of $21$ frames at $T{=}21$, whereas the
bidirectional ConvGRU~\cite{cho2014learning} at $0.17$M and the selective
state-space attacker~\cite{gu2023mamba} at $0.08$M both reach the full clip,
the latter with $5\times$ fewer parameters than the 3D model, so any difference
reflects receptive field rather than capacity.
Table~\ref{tab:redundancy} confirms the ordering holds across all three
architectures while PACE-F's pooling gain is largest under the strongest
attacker, so our figures are conservative.

\subsection{Protocol Controls}

Each control of Sec.~\ref{sec:decomp} changes measured leakage.
\emph{Anchored evaluation} removes a $6.6\times$ swing in evaluation
composition between spans ($T{=}2$ covers $36$ validation videos of $8311$
clips, $T{=}16$ covers $238$ of $896$). \emph{Calibration-free scoring} removes
a second: a fixed-threshold IoU reports $0.0$ for a PACE-T model whose AUC is
$61.2\%$, since ranking is informative while the sigmoid never crosses
the threshold. Fixing optimizer steps and effective batch under gradient
accumulation removes a third. Class frequency also changes gradient-ascent
optimization: SCRUB's unbounded ascent step needs gradient clipping to
converge on the sparser \texttt{car} masks (Tab.~\ref{tab:car} reports the
clipped run), while PACE and PACE-T need no adjustment, since their ascent
enters through the bounded adversarial term of Eq.~\ref{eq:objective}.

\section{Conclusion}
\label{sec:conclusion}

Per-frame verification does not bound what an attacker recovers from a
video segmentation model: averaging a two-second clip recovers the forgotten
class from popular unlearning methods we evaluated on VSPW and most on Virtual KITTI~2,
including models whose per-frame predictions are indistinguishable from a
retrain oracle. A time-shuffled control then identifies redundancy, not
trajectory, as the carrier, across three attacker architectures, two spans,
two classes, and two datasets.

Aiming the objective at that carrier, we built a pooled discriminator that is
permutation-invariant by construction and a two-sided bound closing the frozen-region solution, resulting in PACE: the
lowest pooled leakage of any verdict-safe method that retains usable
utility ($18.79$ AP on VSPW \texttt{person}, $12.10$ on \texttt{car},
$9.11$ on Virtual KITTI~2; retain mIoU $21.2$ and $19.8$ on VSPW, against
at most $8.3$ for comparable baselines).

Three implications follow: video unlearning should be audited with temporal
attackers, using the $\mu_{\mathrm{floor}}{+}2\sigma_{\mathrm{floor}}$ verdict
as a calibration-free pass/fail test against a retrain oracle; defenses
should target the channel the decomposition measures, with the oracle's own
pooling gain as the irreducible floor; and the erasure--utility trade-off is
real: on Virtual KITTI~2 the pooled objective costs $13.5$ mIoU against
PACE-T, making PACE-T the better operating point there. Future work includes
scene-adaptive motion references, longer horizons and moving scenes, where
trajectory may finally carry signal, and extension to subject-level
unlearning and other dense video tasks.

{
    \small
    \bibliographystyle{ieeenat_fullname}
    \bibliography{main}

@String(PAMI = {IEEE Trans. Pattern Anal. Mach. Intell.})

@String(CVPR= {IEEE Conf. Comput. Vis. Pattern Recog.})

@String(ICCV= {Int. Conf. Comput. Vis.})

@String(ECCV= {Eur. Conf. Comput. Vis.})

@String(NIPS= {Adv. Neural Inform. Process. Syst.})

@String(ICLR = {Int. Conf. Learn. Represent.})

@String(AAAI = {AAAI})

@String(ICML  = {ICML})

@String(PAMI  = {IEEE TPAMI})

@String(CVPR  = {CVPR})

@String(ICCV  = {ICCV})

@String(ECCV  = {ECCV})

@String(NIPS  = {NeurIPS})

@String(ICLR  = {ICLR})

@inproceedings{cao2015towards,
  title={Towards making systems forget with machine unlearning},
  author={Cao, Yinzhi and Yang, Junfeng},
  booktitle={IEEE Symp. Security and Privacy}, year={2015}}

@inproceedings{bourtoule2021machine,
  title={Machine unlearning},
  author={Bourtoule, Lucas and Chandrasekaran, Varun and Choquette-Choo, Christopher A. and Jia, Hengrui and Travers, Adelin and Zhang, Baiwu and Lie, David and Papernot, Nicolas},
  booktitle={IEEE Symp. Security and Privacy}, year={2021}}

@inproceedings{golatkar2020eternal,
  title={Eternal sunshine of the spotless net: Selective forgetting in deep networks},
  author={Golatkar, Aditya and Achille, Alessandro and Soatto, Stefano},
  booktitle=CVPR, year={2020}}

@inproceedings{golatkar2020forgetting,
  title={Forgetting outside the box: Scrubbing deep networks of information accessible from input-output observations},
  author={Golatkar, Aditya and Achille, Alessandro and Soatto, Stefano},
  booktitle=ECCV, year={2020}}

@inproceedings{golatkar2021mixed,
  title={Mixed-privacy forgetting in deep networks},
  author={Golatkar, Aditya and Achille, Alessandro and Ravichandran, Avinash and Polito, Marzia and Soatto, Stefano},
  booktitle=CVPR, year={2021}}

@inproceedings{thudi2022unrolling,
  title={Unrolling {SGD}: Understanding factors influencing machine unlearning},
  author={Thudi, Anvith and Deza, Gabriel and Chandrasekaran, Varun and Papernot, Nicolas},
  booktitle={IEEE European Symp. Security and Privacy}, year={2022}}

@inproceedings{fan2024salun,
  title={{SalUn}: Empowering machine unlearning via gradient-based weight saliency in both image classification and generation},
  author={Fan, Chongyu and Liu, Jiancheng and Zhang, Yihua and Wong, Eric and Wei, Dennis and Liu, Sijia},
  booktitle=ICLR, year={2024}}

@inproceedings{foster2024fast,
  title={Fast machine unlearning without retraining through selective synaptic dampening},
  author={Foster, Jack and Schoepf, Stefan and Brintrup, Alexandra},
  booktitle=AAAI, year={2024}}

@inproceedings{kurmanji2023towards,
  title={Towards unbounded machine unlearning},
  author={Kurmanji, Meghdad and Triantafillou, Peter and Hayes, Jamie and Triantafillou, Eleni},
  booktitle=NIPS, year={2023}}

@inproceedings{chundawat2023can,
  title={Can bad teaching induce forgetting? Unlearning in deep networks using an incompetent teacher},
  author={Chundawat, Vikram S. and Tarun, Ayush K. and Mandal, Murari and Kankanhalli, Mohan},
  booktitle=AAAI, year={2023}}

@article{tarun2023fast,
  title={Fast yet effective machine unlearning},
  author={Tarun, Ayush K. and Chundawat, Vikram S. and Mandal, Murari and Kankanhalli, Mohan},
  journal={IEEE Trans. Neural Netw. Learn. Syst.}, year={2023}}

@inproceedings{graves2021amnesiac,
  title={Amnesiac machine learning},
  author={Graves, Laura and Nagisetty, Vineel and Ganesh, Vijay},
  booktitle=AAAI, year={2021}}

@article{nguyen2022survey,
  title={A survey of machine unlearning},
  author={Nguyen, Thanh Tam and Huynh, Thanh Trung and Nguyen, Phi Le and Liew, Alan Wee-Chung and Yin, Hongzhi and Nguyen, Quoc Viet Hung},
  journal={arXiv:2209.02299}, year={2022}}

@inproceedings{triantafillou2024neurips,
  title={Are we making progress in unlearning? Findings from the first {NeurIPS} unlearning competition},
  author={Triantafillou, Eleni and Kairouz, Peter and Pedregosa, Fabian and Hayes, Jamie and Kurmanji, Meghdad and others},
  booktitle=NIPS, year={2024}}

@inproceedings{hayes2024inexact,
  title={Inexact unlearning needs more careful evaluations to avoid a false sense of privacy},
  author={Hayes, Jamie and Shumailov, Ilia and Triantafillou, Eleni and Khalifa, Amr and Papernot, Nicolas},
  booktitle={arXiv:2403.01218}, year={2024}}

@inproceedings{goel2022evaluating,
  title={Towards adversarial evaluation of inexact machine unlearning},
  author={Goel, Shashwat and Prabhu, Ameya and Sanyal, Amartya and Lim, Ser-Nam and Torr, Philip and Kumaraguru, Ponnurangam},
  booktitle={arXiv:2201.06640}, year={2022}}

@inproceedings{cha2024learning,
  title={Learning to unlearn: Instance-wise unlearning for pre-trained classifiers},
  author={Cha, Sungmin and Cho, Sungjun and Hwang, Dasol and Lee, Honglak and Moon, Taesup and Lee, Moontae},
  booktitle=AAAI, year={2024}}

@inproceedings{shokri2017membership,
  title={Membership inference attacks against machine learning models},
  author={Shokri, Reza and Stronati, Marco and Song, Congzheng and Shmatikov, Vitaly},
  booktitle={IEEE Symp. Security and Privacy}, year={2017}}

@inproceedings{carlini2022membership,
  title={Membership inference attacks from first principles},
  author={Carlini, Nicholas and Chien, Steve and Nasr, Milad and Song, Shuang and Terzis, Andreas and Tramer, Florian},
  booktitle={IEEE Symp. Security and Privacy}, year={2022}}

@inproceedings{fredrikson2015model,
  title={Model inversion attacks that exploit confidence information and basic countermeasures},
  author={Fredrikson, Matt and Jha, Somesh and Ristenpart, Thomas},
  booktitle={ACM CCS}, year={2015}}

@inproceedings{zhang2020secret,
  title={The secret revealer: Generative model-inversion attacks against deep neural networks},
  author={Zhang, Yuheng and Jia, Ruoxi and Pei, Hengzhi and Wang, Wenxiao and Li, Bo and Song, Dawn},
  booktitle=CVPR, year={2020}}

@inproceedings{he2019model,
  title={Model inversion attacks against collaborative inference},
  author={He, Zecheng and Zhang, Tianwei and Lee, Ruby B.},
  booktitle={ACSAC}, year={2019}}

@inproceedings{song2019overlearning,
  title={Overlearning reveals sensitive attributes},
  author={Song, Congzheng and Shmatikov, Vitaly},
  booktitle=ICLR, year={2020}}

@inproceedings{goodfellow2014generative,
  title={Generative adversarial nets},
  author={Goodfellow, Ian and Pouget-Abadie, Jean and Mirza, Mehdi and Xu, Bing and Warde-Farley, David and Ozair, Sherjil and Courville, Aaron and Bengio, Yoshua},
  booktitle=NIPS, year={2014}}

@inproceedings{ganin2015unsupervised,
  title={Unsupervised domain adaptation by backpropagation},
  author={Ganin, Yaroslav and Lempitsky, Victor},
  booktitle=ICML, year={2015}}

@inproceedings{edwards2016censoring,
  title={Censoring representations with an adversary},
  author={Edwards, Harrison and Storkey, Amos},
  booktitle=ICLR, year={2016}}

@inproceedings{xie2017controllable,
  title={Controllable invariance through adversarial feature learning},
  author={Xie, Qizhe and Dai, Zihang and Du, Yulun and Hovy, Eduard and Neubig, Graham},
  booktitle=NIPS, year={2017}}

@inproceedings{ren2018learning,
  title={Learning to anonymize faces for privacy preserving action detection},
  author={Ren, Zhongzheng and Lee, Yong Jae and Ryoo, Michael S.},
  booktitle=ECCV, year={2018}}

@inproceedings{wu2020privacy,
  title={Privacy-preserving deep action recognition: An adversarial learning framework and a new dataset},
  author={Wu, Zhenyu and Wang, Haotao and Wang, Zhaowen and Jin, Hailin and Wang, Zhangyang},
  journal=PAMI, year={2020}}

@inproceedings{dave2022spact,
  title={{SPAct}: Self-supervised privacy preservation for action recognition},
  author={Dave, Ishan Rajendrakumar and Chen, Chen and Shah, Mubarak},
  booktitle=CVPR, year={2022}}

@inproceedings{orekondy2018connecting,
  title={Connecting pixels to privacy and utility: Automatic redaction of private information in images},
  author={Orekondy, Tribhuvanesh and Fritz, Mario and Schiele, Bernt},
  booktitle=CVPR, year={2018}}

@inproceedings{miao2021vspw,
  title={{VSPW}: A large-scale dataset for video scene parsing in the wild},
  author={Miao, Jiaxu and Wei, Yunchao and Wu, Yu and Liang, Chen and Li, Guangrui and Yang, Yi},
  booktitle=CVPR, year={2021}}

@article{chen2017deeplab,
  title={{DeepLab}: Semantic image segmentation with deep convolutional nets, atrous convolution, and fully connected {CRFs}},
  author={Chen, Liang-Chieh and Papandreou, George and Kokkinos, Iasonas and Murphy, Kevin and Yuille, Alan L.},
  journal=PAMI, year={2018}}

@article{chen2017rethinking,
  title={Rethinking atrous convolution for semantic image segmentation},
  author={Chen, Liang-Chieh and Papandreou, George and Schroff, Florian and Adam, Hartwig},
  journal={arXiv:1706.05587}, year={2017}}

@inproceedings{long2015fully,
  title={Fully convolutional networks for semantic segmentation},
  author={Long, Jonathan and Shelhamer, Evan and Darrell, Trevor},
  booktitle=CVPR, year={2015}}

@inproceedings{he2016deep,
  title={Deep residual learning for image recognition},
  author={He, Kaiming and Zhang, Xiangyu and Ren, Shaoqing and Sun, Jian},
  booktitle=CVPR, year={2016}}

@inproceedings{zhu2019improving,
  title={Improving semantic segmentation via video propagation and label relaxation},
  author={Zhu, Yi and Sapra, Karan and Reda, Fitsum A. and Shih, Kevin J. and Newsam, Shawn and Tao, Andrew and Catanzaro, Bryan},
  booktitle=CVPR, year={2019}}

@inproceedings{kundu2020video,
  title={Video propagation networks for efficient video semantic segmentation},
  author={Jain, Samvit and Wang, Xin and Gonzalez, Joseph E.},
  booktitle=CVPR, year={2019}}

@inproceedings{pathak2016context,
  title={Context encoders: Feature learning by inpainting},
  author={Pathak, Deepak and Krahenbuhl, Philipp and Donahue, Jeff and Darrell, Trevor and Efros, Alexei A.},
  booktitle=CVPR, year={2016}}

@inproceedings{yu2018generative,
  title={Generative image inpainting with contextual attention},
  author={Yu, Jiahui and Lin, Zhe and Yang, Jimei and Shen, Xiaohui and Lu, Xin and Huang, Thomas S.},
  booktitle=CVPR, year={2018}}

@article{gu2023mamba,
  title={Mamba: Linear-time sequence modeling with selective state spaces},
  author={Gu, Albert and Dao, Tri},
  journal={arXiv:2312.00752}, year={2023}}

@inproceedings{gu2022efficiently,
  title={Efficiently modeling long sequences with structured state spaces},
  author={Gu, Albert and Goel, Karan and R{\'e}, Christopher},
  booktitle=ICLR, year={2022}}

@inproceedings{li2024videomamba,
  title={{VideoMamba}: State space model for efficient video understanding},
  author={Li, Kunchang and Li, Xinhao and Wang, Yi and He, Yinan and Wang, Yali and Wang, Limin and Qiao, Yu},
  booktitle=ECCV, year={2024}}

@inproceedings{cho2014learning,
  title={Learning phrase representations using {RNN} encoder-decoder for statistical machine translation},
  author={Cho, Kyunghyun and Van Merri{\"e}nboer, Bart and Gulcehre, Caglar and Bahdanau, Dzmitry and Bougares, Fethi and Schwenk, Holger and Bengio, Yoshua},
  booktitle={EMNLP}, year={2014}}

@inproceedings{tran2015learning,
  title={Learning spatiotemporal features with 3{D} convolutional networks},
  author={Tran, Du and Bourdev, Lubomir and Fergus, Rob and Torresani, Lorenzo and Paluri, Manohar},
  booktitle=ICCV, year={2015}}

@article{mantelero2013eu,
  title={The {EU} proposal for a general data protection regulation and the roots of the right to be forgotten},
  author={Mantelero, Alessandro},
  journal={Computer Law \& Security Review}, year={2013}}

@inproceedings{hinton2015distilling,
  title={Distilling the knowledge in a neural network},
  author={Hinton, Geoffrey and Vinyals, Oriol and Dean, Jeff},
  booktitle={NIPS Deep Learning Workshop}, year={2015}}

@article{davis2006relationship,
  title={The relationship between {Precision-Recall} and {ROC} curves},
  author={Davis, Jesse and Goadrich, Mark},
  journal={ICML}, year={2006}}

@inproceedings{cabon2020vkitti2,
  title={Virtual {KITTI} 2},
  author={Cabon, Yohann and Murray, Naila and Humenberger, Martin},
  booktitle={arXiv:2001.10773},
  year={2020}}
}

\end{document}